\documentclass[10pt,twocolumn,letterpaper]{article}

\usepackage{cvpr}      

\usepackage{multirow}
\usepackage[linesnumbered,ruled,vlined,boxed,noend]{algorithm2e}

\definecolor{cvprblue}{rgb}{0.21,0.49,0.74}
\usepackage[pagebackref,breaklinks,colorlinks,allcolors=cvprblue]{hyperref}

\usepackage{subfigure}

\def\paperID{4013} 
\def\confName{CVPR}
\def\confYear{2025}

\title{Handling Spatial-Temporal Data Heterogeneity for Federated Continual Learning via Tail Anchor}

\author{Hao Yu\\
Southwestern University of \\Finance and Economics \\
{\tt\small yuhao2033@163.com}
\and
Xin Yang\thanks{Xin Yang is the corresponding author.}\\
Southwestern University of \\Finance and Economics\\
{\tt\small yangxin@swufe.edu.cn}
\and
Le Zhang\\
University of Electronic Science\\ and Technology of China\\
{\tt\small lezhang@uestc.edu.cn }
\and
Hanlin Gu\\
WeBank\\
{\tt\small allengu@webank.com}
\and
Tianrui Li\\
Southwest Jiaotong University\\
{\tt\small trli@swjtu.edu.cn}
\and
Lixin Fan\\
WeBank\\
{\tt\small lixinfan@webank.com}
\and
Qiang Yang\\
Hong Kong University of \\
Science and Technology\\
{\tt\small qyang@cse.ust.hk}
}

\begin{document}
\maketitle
\begin{abstract}
Federated continual learning (FCL) allows each client to continually update its knowledge from task streams, enhancing the applicability of federated learning in real-world scenarios.
However, FCL needs to address not only spatial data heterogeneity between clients but also temporal data heterogeneity between tasks. 
In this paper, empirical experiments demonstrate that such input-level heterogeneity significantly affects the model's internal parameters and outputs, leading to severe spatial-temporal catastrophic forgetting of local and previous knowledge.
To this end, we propose Federated Tail Anchor (FedTA) to mix trainable \textbf{Tail Anchor} with the frozen output features to adjust their position in the feature space, thereby overcoming parameter-forgetting and output-forgetting. 
Three novel components are also included: \textbf{Input Enhancement} for improving the performance of pre-trained models on downstream tasks; \textbf{Selective Input Knowledge Fusion} for fusion of heterogeneous local knowledge on the server; and \textbf{Best Global Prototype Selection} for finding the best anchor point for each class in the feature space.
Extensive experiments demonstrate that FedTA not only outperforms existing FCL methods but also effectively preserves the relative positions of features. 

\end{abstract}    
\section{Introduction}
\label{sec:intro}

\begin{figure*}[htbp]
  \centering
  \includegraphics[width=1\textwidth]{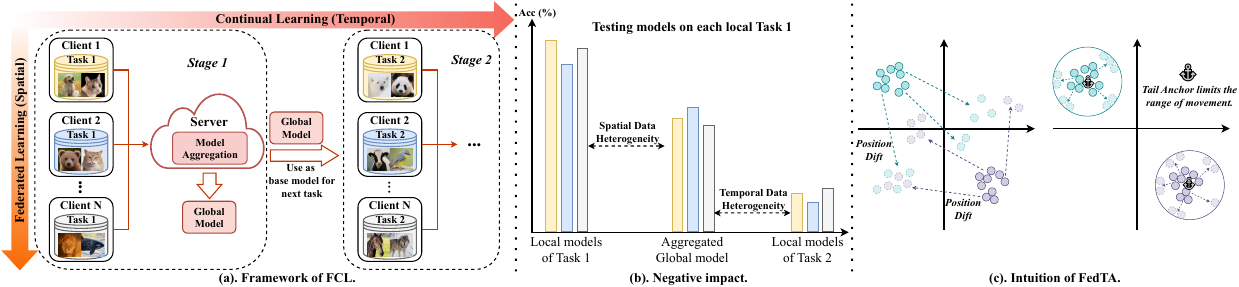}
  \vspace{-5mm}
  \caption{Illustration of FCL, the negative impact of spatial-temporal data heterogeneity and the intuition of Tail Anchor.}
  \label{intro}
  \vspace{-3mm}
\end{figure*}

Data heterogeneity across different clients (Non-IID) is one of the most important challenges in traditional Federated Learning (FL), which greatly hinders the integration of knowledge, leading to the aggregated global model underperforming on local tasks.
Many studies have attempted to address this issue and have made some progress \cite{yang2024fedfed,zhong2023semi,fan2025ten}. However, they are based on an unrealistic static assumption that the training data of all clients will remain unchanged. Federated Continual Learning (FCL) breaks the static limits by allowing clients to continually accumulate knowledge from task sequences \cite{yoon2021federated,babakniya2024data,li2024unleashingpowercontinuallearning,gao2024fedprok}. While FCL expands the applicability of FL in real-world scenarios, it also introduces a more challenging issue, i.e., \textit{spatial-temporal data heterogeneity}. Not only is the data heterogeneous across different clients (spatial), but also the data within different tasks of the same client is heterogeneous (temporal), as shown on the left side of \autoref{intro}.

The most direct negative impact of spatial-temporal data heterogeneity is \textit{spatial-temporal catastrophic forgetting} (ST-CF) \cite{yang2024federatedcl}, as illustrated on the middle side of \autoref{intro}. Catastrophic Forgetting (CF) is a term from Continual Learning (CL), used to describe the phenomenon where a deep model, after learning multiple tasks, tends to forget the knowledge of previous tasks, resulting in a decrease in accuracy \cite{kirkpatrick2017overcoming,hadsell2020embracing,li2017learning}. 
In FCL, clients face \textit{temporal catastrophic forgetting} as local models would continually learn different tasks over time. 
Additionally, Non-IID data would lead the aggregated global model to \textit{spatial catastrophic forgetting} (i.e., a decline in the performance of the global model on local test sets). 
Moreover, \textit{spatial forgetting will interact with temporal forgetting}, as clients use the global model as the base model to learn the next task \cite{yang2024federatedcl}.


Spatial-temporal data heterogeneity, which manifests as differences at the \textit{model input}, leads to corresponding changes in \textit{model parameters and outputs}.
We believe this is the fundamental cause of forgetting. 
To be specific, as data changes over time and space, feature extractor and classification head of the model will adapt to the most recent inputs, leading to forgetting of previous and client-specific knowledge (a detailed analysis of the effect of forgetting is presented in Fig. \ref{intro}(a-b) and \autoref{preliminary_experiment}).
Fortunately, the use of pre-trained large models can effectively mitigate catastrophic forgetting at the parameter level, as they have sufficient capacity to extract features without changing internal parameters \cite{wang2022continual,smith2023closer,ermis2022continual}. However, frozen pre-trained models often perform poorly in downstream tasks, making them unsuitable for direct application~\cite{kang2023grounding}. Additionally, they cannot handle forgetting at the output since the classification head is trainable and will fit to the most recent task data.


To fully leverage the power of pre-trained models and address spatial-temporal forgetting from both the parameter and the output aspects, we first define Tail Anchors and mix them with frozen output features to fix the position, as shown in Fig. \ref{intro}(c).
Based on this concept, we propose \textbf{Fed}erated \textbf{T}ail \textbf{A}nchor (FedTA), which leverages the tail anchor to keep the positions of each class invariant.

Firstly, each client shares a pre-trained Vision Transformer (ViT) \cite{dosovitskiy2020image} and a two-stage training strategy is designed to enhance the performance of pre-trained models at the input level and alleviate forgetting at the output level. By adding \textit{tail anchors} to the output features, the features of samples that experience forgetting can quickly return to their original positions in the feature space, thereby avoiding forgetting caused by spatial-temporal changes. After completing local training, the server will separately process the parameters added during input enhancement by the client and the local prototype. On the one hand, we design a \textit{selective input knowledge fusion} mechanism to selectively integrate the knowledge used for input enhancement from different clients; on the other hand, the server will calculate the similarity between local prototypes to form a similarity adjacency matrix. In each iteration, the server will select the local prototype with the lowest average similarity for each class as the global prototype. 
If the average similarity falls below a threshold, the global prototype will be fixed to prevent forgetting. 

Currently, there is little research on spatial-temporal heterogeneity in FCL \cite{fan2025ten}. To our knowledge, we are the first to attempt to address both temporal and spatial data heterogeneity from the perspective of forgetting. The main contributions can be summarized as:

\begin{enumerate}
\item  Empirical experiments are conducted to show that spatial-temporal data heterogeneity can cause significant changes in the important features extracted by the model for the same samples, and it also causes shifts in their positions within the feature space. This leads to severe ST-CF of previous knowledge and local knowledge. We refer to these two phenomena as ``parameter-forgetting" and ``output-forgetting", respectively.

\item FedTA leverages a pre-trained ViT along with four novel components, aiming to prevent features from shifting their positions in the feature space due to spatial-temporal data heterogeneity. FedTA efficiently overcomes both parameter-forgetting and output-forgetting in FCL caused by spatial-temporal data heterogeneity.

\item Extensive experiments not only demonstrate the state-of-the-art performance of FedTA but also show its remarkable ability to resist spatial-temporal forgetting. Moreover, visualization results indicate that FedTA effectively preserves the relative positions of features, preventing positional shifts due to spatial-temporal variations.

\end{enumerate}

\section{Related Work}

Spatial data heterogeneity, as known as the Non-IID problem, has attracted much attention \cite{kim2023navigating, pieri2023handling, crawshaw2024federated, zhao2018federated}. Existing methods tackle data heterogeneity by either incorporating more effective local training or devising more comprehensive aggregation mechanisms \cite{li2022federated}.

Although these studies have made progress in overcoming spatial data heterogeneity, they are unable to cope with more realistic and dynamic scenarios where each client continually learns on their own task stream.

FCL has indeed greatly enhanced the practical value of FL in real-world scenarios, especially on the edge computing side \cite{zhang2022cross, xu2023age,zhong2022flee}. It allows each client to rapidly learn knowledge from the current task without forgetting previously knowledge, thus avoiding the need to retrain from scratch and greatly saving computational resources.


In a survey paper on FCL, the authors identified a key issue that existing FCL articles have overlooked: the interaction between spatial heterogeneity and temporal heterogeneity, which leads to a unique challenge: spatial-temporal catastrophic forgetting (ST-CF) \cite{yang2024federatedcl}. It means that models not only forget previous knowledge due to continual learning but also forget local knowledge due to federated aggregation. Existing FCL methods do not realize that spatial heterogeneity can exacerbate the temporal forgetting, so when the spatial heterogeneity becomes stronger, the performance is not as expected. Besides, effective FCL methods currently heavily rely on replaying or generating pseudo data to mitigate the effects of spatial-temporal data heterogeneity \cite{zhang2023target,babakniya2024data,liang2025diffusion,li2024sr,li2024towards,li2025personalized}. However, this may pose certain privacy risks and incur high computational cost.

Only a very small portion of work has attempted to address data heterogeneity from time and space simultaneously now \cite{yu2024overcoming,li2024facing}. However, none of them have delved into how this heterogeneity leads to forgetting, nor have they ensured sufficiently strong spatial-temporal heterogeneity in their experimental settings. To our best knowledge, we are the first to deeply analyze how the heterogeneity of inputs affects model parameters and outputs. 
\section{Spatial-Temporal Data Heterogeneity}
\label{sec:2}

\subsection{Problem Definition}
The purpose of Spatial-Temporal Data Heterogeneity is to continually integrate knowledge from different clients and different time periods. We extend the traditional FL to FCL with strong spatial-temporal data heterogeneity. 

For spatial heterogeneity, given $a$ clients (denoted as $\mathcal{A}=\{A_1, A_2, \ldots, A_a\}$), and a central server $S$, each client's data is composed of private classes $C_v^i$ and public classes $C_p$, where private classes refer to the class of data that can only be seen by the client itself. We ensure that the data of $C_p$ is non-overlapping between clients. Further, we can set $|C_p|=0$ to ensure extreme spatial heterogeneity. We run experiments with $|C_p|=0$ on Imagenet-R dataset.

For temporal heterogeneity, the task sequence of client $A_i$ is denoted as $\mathcal{T}_i=\{T_i^1, T_i^2, \ldots, T_i^{n_i}\}$, where $n_i$ represents the total number of tasks on client $A_i$. Each task consists of the same number but entirely different classes. 
  
During the training of task $r$, the global model on the server already possesses the knowledge of $T_i^1$ to $T_i^{r-1}$ from client $\{A_i, 1 \leq i \leq a\}$. The server $S$ then distributes it back to clients. After personalizing the received global model $\theta_g^{r-1}$, the client $A_i$ continually trains it on $T_i^r$ as the initial model to get the new local model $\theta_i^r$. The local model $\theta_i^r$ should perform well in classifying classes of $\{\mathcal{T}_1^{1} \cup \mathcal{T}_1^{r-1} \ldots, \mathcal{T}_i^{1} \cup \mathcal{T}_i^r\ \cup \ldots, \mathcal{T}_a^{1} \cup \mathcal{T}_a^{r-1}\}$.

\subsection{Negative Impact}
\label{preliminary_experiment}

Spatial-temporal data heterogeneity is a type of heterogeneity in model inputs. Due to the back-propagation mechanism, it would significantly affect the internal parameters of the model and the outputs \cite{rumelhart1986learning}. It not only introduces differences between models of different clients but also causes the features output by the same sample to undergo significant changes, thereby causing spatial-temporal forgetting of previous knowledge and local knowledge.

\begin{figure*}[htbp]
  \centering
  \includegraphics[width=0.8\textwidth]{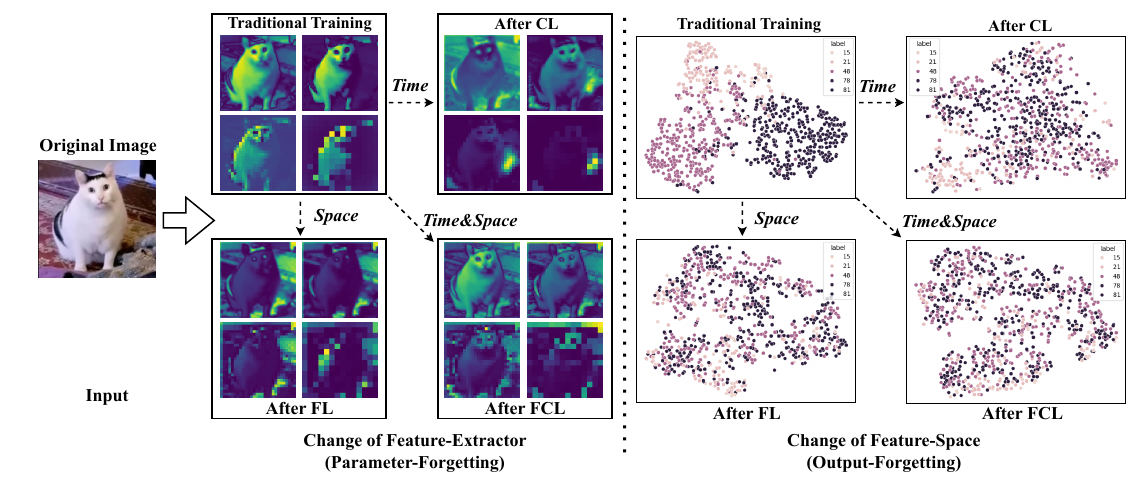}
  \vspace{-4mm}
  \caption{Illustrations of the negative impact of spatial-temporal data heterogeneity on the feature extractor and feature space. [Left Side] illustrates the variation of significant features extracted by the feature extractor for the same input sample, where brighter colors indicate more important features. As spatial-temporal changes occur, the extracted features gradually shift away from ``cat", even extracting features near the image edges. [Right Side] depicts the changes in the positions of the features in the feature space after undergoing spatial-temporal transformations for the same batch. }
  \label{impact}
  \vspace{-4mm}
\end{figure*}

\textbf{For the feature extractor}, as shown on \textit{the left side} of \autoref{impact}, traditional centralized single-task training can accurately extract beneficial features. However, after continual learning of four other tasks, for the same image, the extracted features are completely unrelated to the cat. Similarly, for FL, the aggregated global model also fails to extract features related to the cat itself. More critically, when the model encounters spatial-temporal data heterogeneity, the crucial features extracted by the feature extraction layer are predominantly concentrated at the edges of the images, which is highly negative to classification task. We term this phenomenon as \textbf{parameter-forgetting}.

\textbf{For the output (feature space)}, as shown on \textit{the right side} of \autoref{impact}, the features extracted by the initial model have clear classification boundaries, and the features of each class are relatively concentrated in the same area. However, after CL or FL, the features extracted for the same samples no longer possess clear boundaries, especially in FCL. Moreover, the positions of the features gradually deviate from their original locations with the spatial-temporal variation, leading to the forgetting of old samples. We term this phenomenon as \textbf{output-forgetting}.

Let's delve even further into the effects of spatial-temporal data heterogeneity. For deep neural networks, changes at the input level directly affect model parameters and corresponding outputs, \textit{\textbf{thereby causing continual variations of the feature space.}} For spatial data heterogeneity, the absence of a common feature space among clients makes it challenging to share heterogeneous knowledge. For temporal data heterogeneity, changes in the feature space over time lead to variations in the locations of features of the same samples. If we can address the issues mentioned above simultaneously, then spatial-temporal catastrophic forgetting will be resolved.

\subsection{Motivation}
It is evident that spatial-temporal data heterogeneity leads to both parameter-forgetting and output-forgetting. Therefore, methods need to possess the following three capabilities: (1) Ensure that the model extracts nearly identical features for the same sample; (2) Fix the positions of extracted features. (3) Allow clients to have a common feature space to better utilize heterogeneous knowledge.

However, due to the training method of deep networks and the large number of parameters involved, parameter updates are uncontrollable, \textit{making it nearly impossible to mitigate parameter-forgetting}. Similarly, since ensuring consistency within the parameters is impossible, it is also hard to guarantee the invariance of outputs in the feature space. 

Pre-trained large models have attracted attention due to their powerful representation capabilities. There are already articles attempting to apply pre-trained ViT to overcome forgetting in CL \cite{li2024learning,pelosin2022towards}. Inspired by this, we find that freezing the feature extractor of pre-trained ViT can effectively \textbf{eliminate parameter-forgetting}. In FCL, clients share the same pre-trained model, ensuring that they have \textbf{the same knowledge/feature space}, which makes knowledge transfer between clients easier. Furthermore, by mixing learnable parameters (referred to ``tail anchor" in this paper) with frozen features, we can effectively control their positions in the feature space, thus \textbf{addressing output-forgetting}. The server selects anchor points with the lowest similarity to other classes’ anchor points as the global anchor points in the feature space. During client-side training, the tail anchor converges towards these class-specific anchor points. Therefore, we mitigate the performance degradation caused by parameter-forgetting and output-forgetting.

\section{Methodology: FedTA}
To address the spatial-temporal catastrophic forgetting, which includes parameter-forgetting in the feature extractor and output-forgetting in the feature space, we propose FedTA. Its aim is to leverage a frozen pre-trained model and cross-mix learnable parameters after the output features, ensuring that the position of features in the feature space remains fixed and unaffected by spatial-temporal changes.



\begin{figure*}[htbp]
  \centering
  \includegraphics[width=0.8\textwidth]{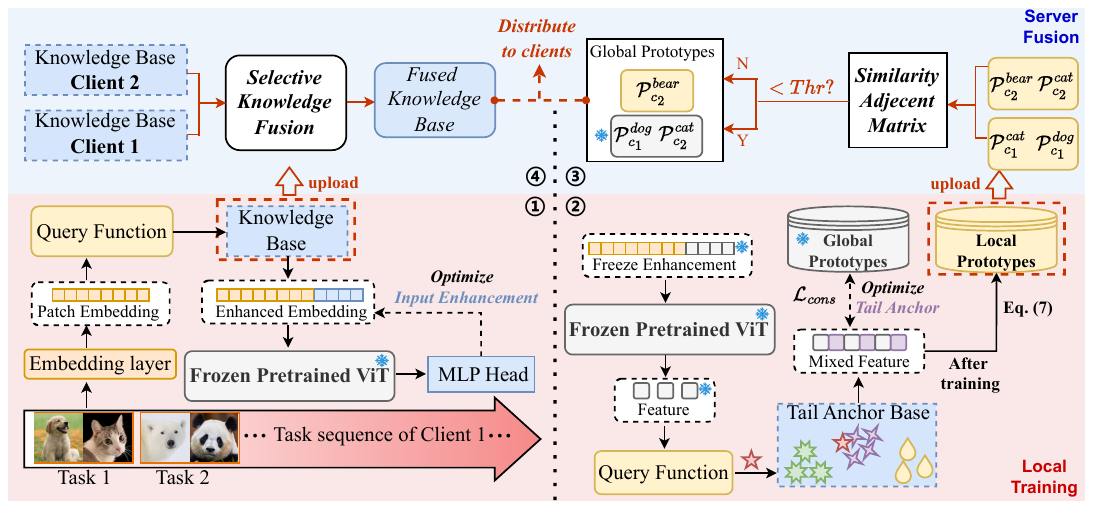}
  \caption{An overview of FedTA. Local training is a two-stage training process. The first stage involves adding \textbf{\textit{input enhancement}} to the image embeddings to fully utilize ViT (see \textbf{1}). In the second stage, the extracted features are fixed, and the corresponding \textbf{\textit{tail anchor}} is mixed with them to adjust the similarity between classes by applying contrastive learning with global prototypes (see \textbf{2}). Then, the local knowledge base of input enhancements and the local prototypes of each class are uploaded to the server, where \textbf{\textit{selective input knowledge fusion}} for the knowledge base (see \textbf{4}) and \textbf{\textit{global best prototype selection}} for the local prototypes (see \textbf{3}) are performed, respectively.}
  \vspace{-3mm}
  \label{method}
\end{figure*}

\subsection{Input Enhancement}
\label{sec4.1}
We assign each client with the same pre-trained ViT model as a foundation model. With its parameters frozen, clients learn common knowledge that operate at the input level. The purpose is to extract knowledge into a common space through the same model and enhance ViT's performance.

\noindent\textbf{Knowledge Base.} We devise a knowledge base for storing and selecting the input enhancement parameters. The knowledge base of client $i$ is defined as 
\begin{equation}
    \mathcal{KB}_{i}=\{IE_1, IE_2, \ldots, IE_M\},
\end{equation}
where $M$ is the base size and $IE$ is a set of learnable parameters. Then, let $x$ and $E = f_e(x)$ be the input and its corresponding embedding feature, respectively. $f_e(\cdot)$ refers to the embedding function of ViT. Denoting $\{s_i\}_1^N$ be the indices of selected $N$ sets, then we can modify the embedding feature as follows:
\begin{equation}
    E' = \left[ IE_{s_1}, \ldots, IE_{s_N} ; E\right], 1 \leq N \leq M,
\end{equation}
where [;] represents concatenation along the token length dimension. Each $IE$ has a corresponding key, denoted as $K^{ie}$, to facilitate the selection of the $IE$ based on the similarity of keys.

\noindent\textbf{Optimization for the input enhancement.} Each client has a classification head used for training input enhancement parameters, denoted as $H^i_e$. At the beginning of training, it is necessary to load the pre-trained model with $H^i_e$ to enable it to perform the classification task, and we denote the model with $H^i_e$ as $\mathcal{V}^i_e$. Overall, the training loss function is as follows:
\begin{equation}
\label{loss1}
    \operatorname{min} \mathcal{L}(\mathcal{V}^i_e(E'),y) + \lambda_1 \sum_{\mathcal{K}^{ie}_{s}} \operatorname{dis}(K^{ie}_{in},K{s_i}^{ie}),
\end{equation}
where $\lambda_1$ is a hyperparameter, $K^{ie}_s$ and $K^{ie}_{in}$ are used to find the best $IE$. The initial term comprises the softmax cross-entropy loss to optimize selected $IE$s, while the subsequent term serves as a surrogate loss aimed at bringing selected keys closer to their corresponding query features. Cosine similarity is used as the distance function.

\textbf{In a nutshell}, $IE$ is a set of learnable parameters that can be concatenated to the image embedding $E$ to enhance the performance of ViT. Each $IE$ has a corresponding key $K^{ie}$. $E$ is first processed by ViT to obtain features, which are then used as a query key $K^{ie}_{in}$ to calculate the similarity with the key of each $IE$. Then best $IE_s$ is selected to form a \textbf{enhance embedding $E'$}.

\subsection{Tail Anchor}
\label{sec4.2}
\noindent\textbf{Query function.} Once the input enhancement parameters are well trained, they will be frozen, including their corresponding keys, until the next task training. The enhanced input embedding $E'$ would be processed by the frozen ViT \cite{dosovitskiy2020image} again to get the features, denoted as $\mathcal{F}_{out}$. Then it will be used as the key to find the corresponding tail anchor based on the cosine similarity. 

Tail Anchor is defined as key-value pairs for $m$ classes: $\mathcal{TA} = \{(K_1^{ta}, TA_1), (K_2^{ta}, TA_2), \ldots, (K_m^{ta}, TA_m)\}$. Specifically, $K^{ta}_s, s \in [m]$ is obtained as:
\begin{equation}
    K^{ta}_s = \underset{\mathcal{K}^{ta}}{\operatorname{argmin}} \operatorname{dis}(\mathcal{F}_{out}, K^{ta}_i),
\end{equation}
where $K^{ta}_s$ denotes the chosen tail anchor's key, and $\mathcal{K}^{ta}$ represents the set of keys for all tail anchors. 

\textbf{In a nutshell,} the tail anchor of each class acts as additional parameters to manage the distance to a fixed position (global prototype of each class) in the feature space. Its main advantage is its relative position, which remains consistent regardless of changes in space and time. This consistency ensures that the tail anchor can effectively guide the positioning of general output features.


\noindent\textbf{Optimization for the tail anchor.} Once the tail anchor is chosen, it will be mixed with $\mathcal{F}_{out}$ to form a new feature $\mathcal{F}_{TA}$. If a client has global prototypes (i.e., not the first round), then contrastive learning is utilized to unify the features across clients through the following unified representation loss function:
\begin{equation}
\label{loss2}
    \mathcal{L}_{cons}\left(\mathcal{F}_{TA}\right)=-\log \frac{\exp \left(\mathcal{F}_{TA} \cdot \mathcal{G}^{y} / \tau\right)}{\sum_{y_{a} \in \mathcal{Y}^{t}} \exp \left(\mathcal{F}_{TA} \cdot \mathcal{G}^{\left.y_{a} / \tau\right)}\right.},
\end{equation}
where $\mathcal{Y}^t$ represents the global available classes up to task $t$ and $\mathcal{G}^y$ represents the global prototypes of class $y$. $\tau$ denotes the temperature that controls the tolerance of difference between extracted features and the corresponding global prototype. The overall loss function to optimize the tail anchor can be formulated as follows:
\begin{equation}
    \mathcal{L}_{ta} = \mathcal{L}_{CE} (\mathcal{F}_{TA}) + \lambda_2 \mathcal{L}_{cons}\left(\mathcal{F}_{TA}\right) + \lambda_3 \operatorname{dis}(\mathcal{F}_{TA},K^{ta}_s),
\end{equation}
where $\mathcal{L}_{CE}$ is the standard cross-entropy loss, the second term is to adjust its position in the feature space through contrastive loss with the global prototypes (fixed positions). The last term is to enhance the similarity between the selected key and the query key.

\noindent\textbf{Local prototypes.} Once the training process of the tail anchor is done, the tail anchors will be frozen and remain unchanged. The local prototype is obtained by averaging features with tail anchors belonging to the same class, computed through
\begin{equation}
    P_{i}^{y}=\frac{1}{\left|\mathcal{D}_{a}^{y}\right|} \sum_{(x, y) \in \mathcal{D}_{a}^{y}} \mathcal{F}_{TA}^{x},
\end{equation}
where $\mathcal{D}_{a}^{y}$ denotes the subset of private dataset of client $a$ of class $y$. Each client forms a local set of prototypes, which is then uploaded to the server. The server iteratively selects the prototype with the lowest average similarity as the global prototype for that class.

\subsection{Selective Input Knowledge Fusion}
\label{sec4.3}

We follow a common setting, which allows the server to possess a small-scale surrogate dataset, denoted as $\mathcal{D}_s$. $\{x_s,y_s\}$ are the samples and corresponding labels from $\mathcal{D}_s$ for the distillation process. 
Assuming the total number of $\mathcal{KB}$ is $n$, $\mathcal{KB}_i$ is randomly selected as the target for distillation. $E'_i$ represents the enhanced embedding formed by concatenating the $x_s$'s embedding with the $IE$ from $\mathcal{KB}_i$, where $\mathcal{V}(\cdot)$ denotes the ViT's feature extraction process. Therefore, the distillation loss can be formulated as follows:
\begin{equation} 
\mathcal{L}_{KD} = \frac{1}{n-1} \sum_{\substack{j=1 \\ j \neq i}}^n \underset{x_s \in \mathcal{D}_s}{\operatorname{MSE}}(\mathcal{V}(E'_i),\mathcal{V}(E'_j)). 
\end{equation}


\subsection{Best Global Prototype Selection}
\label{sec4.4}
When the server receives local prototype sets from different clients, it reorders them to form a new set $P_G$ according to the class. Specifically, when two clients both have prototypes related to class $q$, denoted as $P_i^q$ and $P_j^q$, they will be adjacent to each other in the reordered prototype set. Then, the server computes the similarity between each pair of sets in the collection, forming an adjacency matrix $\mathcal{M}$. The element of $\mathcal{M}$ is computed through:
\begin{equation}
    \mathcal{M}_{ij} = \operatorname{dis}(P_G^i, P_G^j), 0<i\leq j\leq \left| P_G \right |.
\end{equation}
Notice that if $P_G^i$ and $P_G^j$ belong to same class, then $\mathcal{M}_{ij} = 1$. In each round, the server selects the prototype with the lowest average similarity with all local prototypes as the global prototype for one class. The selection process of the global prototype $\mathcal{G}^y$ for class $y$ can be expressed as follows:
\begin{equation}
    \mathcal{G}^y = P_g^s = \underset{y_{low} \leq i \leq y_{high}}{\operatorname{argmin}} \bar{\mathcal{M}_i} = \frac{1}{\left| P_G \right |} \sum_{j=1}^{\left| P_G \right |} A_{ij},
\end{equation}
where $y_{low}$ and $y_{high}$ are the start index and end index of the local prototypes of class $y$ in $P_G$. $P_g^s$ is the local prototype who has lowest similarity for class $y$. If the average similarity $\bar{\mathcal{M}_i}$ falls below the threshold $Thr$ during the iteration process, then that prototype is fixed as the global prototype for its class and will not be altered further. As a result, this global prototype will serve as a fixed anchor point for that class in the feature space.

\section{Experiments}
\subsection{Setup}
\begin{table*}[!h]
\small
\begin{center}
\caption{Accuracy of the aggregated global model on local test sets with 5 class-incremental tasks.}
\label{tab:cifar100}
\begin{tabular}{c|c|ccccc|ccccc}
\toprule
\multirow{2}{*}{Algorithm}  & \multirow{2}{*}{Type} & \multicolumn{5}{c}{CIFAR-100 Task ID} & \multicolumn{5}{c}{Imagenet-R Task ID} \\

 & &  1& 2 & 3 & 4 & 5 & 1& 2 & 3 & 4 & 5 \\

\midrule
FedAvg \cite{mcmahan2017communication}  & \multirow{3}{*}{FL} & 43.9 &50.6 & 57.3 & 55.5 & 61.2  & 37.7 &35.4 & 35.5 &35.8 &36.7    \\

FedProx \cite{li2020federated}  &  &23.7 & 22.8 & 26.0 & 22.1 & 23.6 & 20.2 & 19.7 &19.7 & 18.9 & 17.8 \\

FedNova \cite{wang2020tackling}& &13.7 & 18.8 & 20.1 & 16.1 &15.1 & 2.0 & 4.7 &4.6 & 7.8 &7.7 \\

\midrule

FedLwF \cite{li2017learning} & \multirow{4}{*}{FL+CL} & 36.9& 12.5 & 17.1 & 13.6 & 9.7 &5.9 & 7.0 &2.6 &4.0 & 3.8 
 \\

FedViT \cite{dosovitskiy2020image}&  & 70.2 & 70.0 &71.4 & 66.0 & 67.3 & 68.2 & 59.8 & 57.3 & 59.8 & 57.9 \\ 

FedL2P \cite{wang2022continual}  & &28.4 & 29.9 & 29.3 & 25.4 &25.7 & 27.1 & 27.6 & 24.8 & 25.1 & 26.5\\ 

FedDualP \cite{wang2022dualprompt} &  &31.7 & 42.8 & 52.8 & 39.0 & 46.3 & 23.5 &26.6 &26.4 & 30.2 & 32.0 \\
\midrule

GLFC \cite{dong2022federated} & \multirow{4}{*}{FCL} &82.0 & 63.1 & 73.4 & 64.2 & 64.8 &	61.9 & 67.1 & 67.0 & 71.7 & 57.2 
 \\

TARGET \cite{zhang2023target}& & 54.0  & 41.4 & 32.2 & 13.9 & 15.9 
 & 39.9 & 15.0  & 16.0 &  17.5 &16.1 \\ 

MFCL \cite{babakniya2024data} & & 46.7 & 16.2 & 10.6 & 14.6 & 13.5& 28.8 &14.5&16.2 & 13.3 & 13.8 \\

FedMGP \cite{yu2024personalized} &  & 90.2 & 85.3 &90.7 & 89.2 & 82.2 & 77.3 & 76.8 &78.0 & 75.6  & 75.4 \\

\midrule

Ours (FedTA) & \multirow{4}{*}{FCL} & \textbf{96.1} & \textbf{94.0}	& \textbf{94.6 } & \textbf{94.4 } & \textbf{89.4}  & \textbf{81.5 } & 78.8  & 79.2 & 80.6 & \textbf{85.0}\\

Ours-w/o TA &  &78.7 & 75.5 &	73.4 &	69.3 &	72.3 & 79.6 & 72.3 &	72.3 &	74.1 & 63.2 \\

Ours-w/o SIKF &  & 90.7 & 88.8 & 91.1 & 91.4  & 89.1 & 80.0 & \textbf{80.5} & \textbf{81.1} & \textbf{82.9} & 81.7 \\

Ours-w/o BGPS & & 90.8 & 92.5 & 88.4 & 91.4 & 88.6 & 80.5 & 78.7 & 78.3 & 81.7 & 79.1 
 
\\

\bottomrule
\end{tabular}
\end{center}
\vspace{-7mm}
\end{table*}

\noindent\textbf{Continual Setting.} To ensure that the impact of spatial-temporal data heterogeneity is adequately reflected, we partition the data as follows: Each client continually learn from a task sequence of 5 tasks, and there are 5 clients in the experimental setting. For \textit{CIFAR-100} \cite{krizhevsky2009learning}, we allow each client to have access to 15 private classes exclusive to itself, resulting in 25 public classes. Thus, each client has data for a total of 40 classes, with each task consisting of only 8 classes. For \textit{ImageNet-R} \cite{hendrycks2021many}, we exacerbate spatial data heterogeneity by assigning 40 private classes to each client, with no public classes across clients. Similarly, each task consists of 8 classes. Notice that we use the Dirichlet distribution for the public classes to assign data, ensuring there is no overlap between different clients.

\noindent\textbf{Surrogate Data.} We follow the common setting \cite{ma2022continual,huang2022learn} where the server possesses a small surrogate dataset. For CIFAR-100, each class has only 20 samples, while for ImageNet-R, each class has only 5 samples.


\subsection{Metrics}
To verify whether the method can effectively address the challenges brought by spatial-temporal data heterogeneity, we use two new metrics from \cite{yang2024federatedcl} to evaluate the performance of mitigating forgetting. 

\textit{Definition 1. \textbf{(Temporal Knowledge Retention)}}:

\begin{equation}
\label{krt}
    KR_t=\frac{1}{a}\sum_{i=1}^a\frac{Acc(\theta^r_i;T^0_i)}{Acc(\theta^0_i;T^0_i)},
\end{equation}
where $Acc(\theta^r_i;T^0_i)$ denotes the test accuracy of client $A_i$'s local model at $r$-th round on the $0$-th task and $Acc(\theta^0_i;T^0_i)$ denotes the accuracy of client $A_i$'s local model at the initial round on the $0$-th task.

\textit{Definition 2. \textbf{(Spatial Knowledge Retention)}}:
\begin{equation}
\label{krs}
    KR_s=\frac{1}{a}\sum_{i=1}^a\frac{Acc(\theta^r_g;T^r_i)}{Acc(\theta^r_i;T^r_i)},
\end{equation}
where $Acc(\theta^r_g;T^r_i)$ denotes the accuracy of the global model $\theta^r_g$ on the current local task $T^r_i$ at client $A_i$ and $Acc(\theta^r_i;T^r_i)$ denotes the accuracy of the local model $\theta^r_i$ on its current local task $T^r_i$.

\subsection{Results \& Ablation Study}

\begin{figure}[h]
    \centering   
    \subfigure[$KR_s$ on CIFAR-100.] 
        {
		\begin{minipage}[c]{.45\linewidth} 
				\centering
				\includegraphics[scale=0.28]{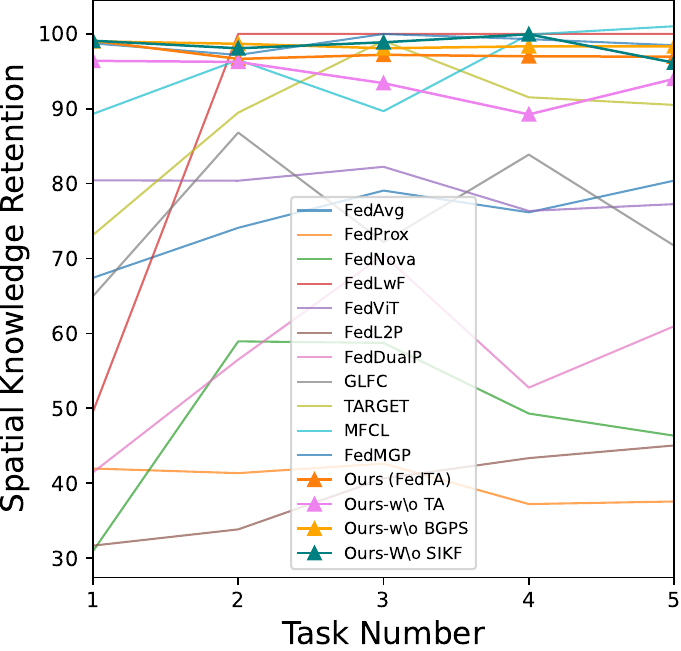}
				\end{minipage}
                \label{fig_akrs}
			}
		\subfigure[$KR_t$ on CIFAR-100.]
		{
			\begin{minipage}[c]{.45\linewidth}
				\centering
				\includegraphics[scale=0.28]{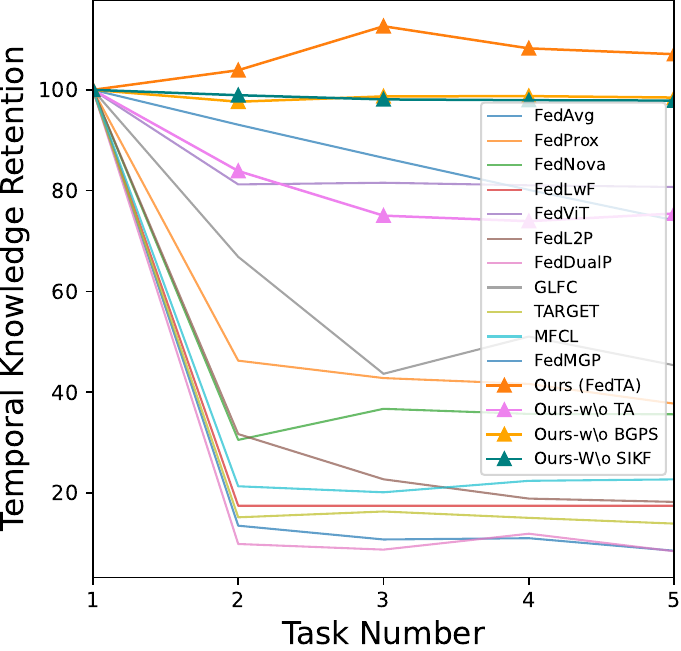}
				\end{minipage}
                \label{fig_akrt}
		}
  
  \subfigure[$KR_s$ on ImageNet-R.]
		{
			\begin{minipage}[b]{.45\linewidth}
				\centering
				\includegraphics[scale=0.28]{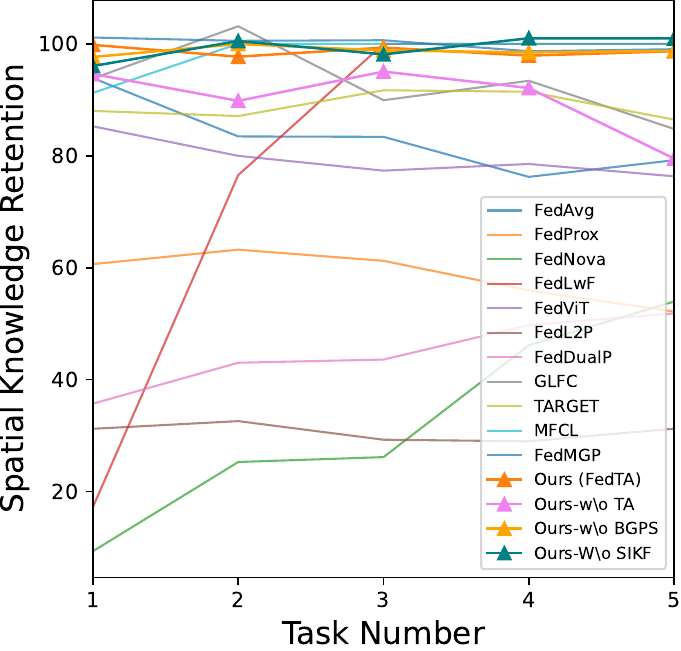}
				\end{minipage}
                \label{fig_skrs}
		}
    \subfigure[$KR_t$ on ImageNet-R.]
		{
			\begin{minipage}[b]{.45\linewidth}
				\centering
				\includegraphics[scale=0.28]{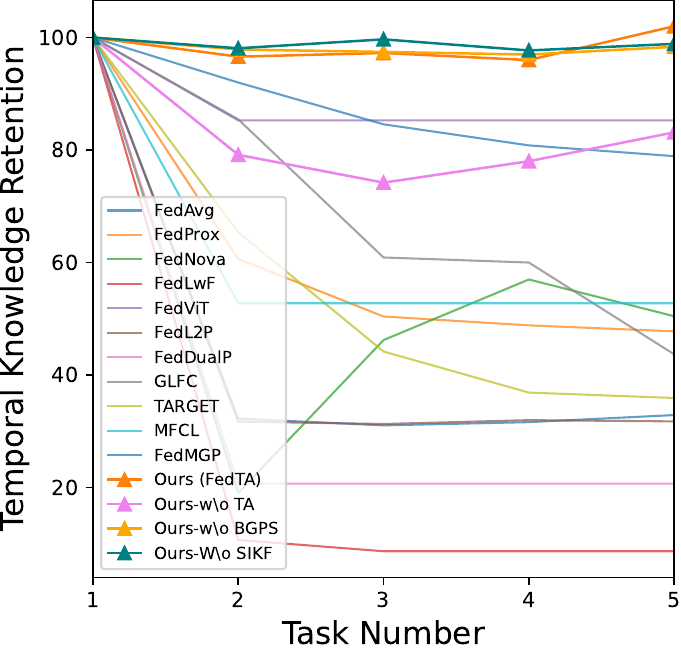}
				\end{minipage}
                \label{fig_skrt}
		}
   \vspace{-4mm}
   \caption{Knowledge retention on different dataset.}
   \vspace{-4mm}
   \label{ablation}
\end{figure}

\begin{figure*}[!h]
  \centering
  \includegraphics[width=0.8\textwidth]{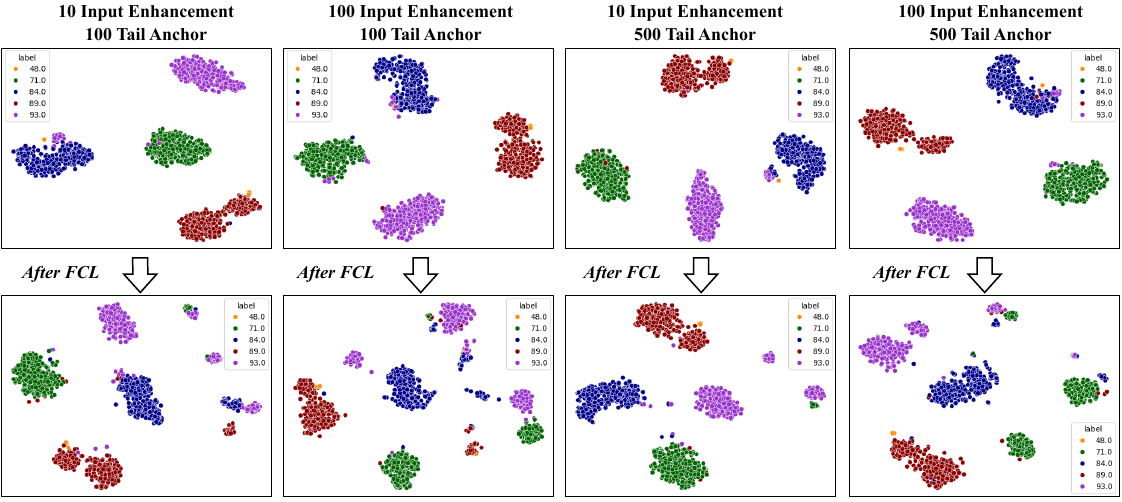}
  \vspace{-3mm}
  \caption{T-SNE for position changes of features corresponding to the same samples after FCL.}
  \vspace{-4mm}
  \label{sensitivity analysis}
\end{figure*}

\autoref{tab:cifar100} illustrates the average accuracy of the aggregated global model on local test sets. The performance of \textit{FedViT} is acceptable because all the parameters of its feature extractor are frozen, and only the classification head is involved in training and aggregation. However, it still experiences a certain degree of forgetting. \textit{FedL2P} and \textit{FedDualP}, which introduce trainable parameters on the input side and within the model, perform very well on the local side, achieving around 90\% accuracy. However, as we concluded in \autoref{preliminary_experiment}, almost all trainable parameters are directly affected by the data. Consequently, after aggregation, there is significant forgetting on the local test sets. 

Surprisingly, the performance of \textit{TARGET}, \textit{FedLwF} and \textit{MFCL}, the three baseline methods that use replay data to mitigate forgetting, is extremely poor. We speculate that the large size of data ($3\times224\times224$) results in the low quality of replayed pseudo-samples. Moreover, replay-based methods pose a certain risk of privacy leakage in federated learning, limiting the further application of these methods in real-world scenarios. \textit{GLFC} also suffers significant performance degradation when faced with severe spatial-temporal data heterogeneity. However, its performance in \autoref{tab:cifar100} remains the best among the baseline methods.

\textit{FedTA} demonstrates the superior performance in these two settings, indicating its successful mitigation of the impact of spatial data heterogeneity. Furthermore, ablation studies highlight the effectiveness of the proposed novel components, with the \textit{Tail Anchor} contributing the most to the performance improvement. However, the selective input knowledge fusion at the server-side sometimes falls below the results of direct weighted averaging on ImageNet-R. We believe this is due to insufficient surrogate data, which prevents adequate selective knowledge fusion.

\autoref{ablation} further illustrates the impact of spatial-temporal data heterogeneity on the methods using \autoref{krs} and \autoref{krt}. Only FedTA maintains a high level of temporal and spatial knowledge retention, both at around 98\%. While other baseline methods, especially in $KR_t$, perform extremely poorly. \textit{Tail Anchor} also has been verified to play a significant role in overcoming ST-CF. 

\noindent\textbf{Visualization \& Sensitivity Analysis.}
\autoref{sensitivity analysis} illustrates that FedTA can effectively control the relative distances between features. 


\subsection{Privacy \& Efficiency Analysis}

\noindent\textbf{Computational Burden.} During the local training phase, clients train both the Input Enhancement and Tail Anchor components while the ViT itself remains frozen. Therefore, the number of parameters in these two components, along with the classification head, determines the training overhead of FedTA. The size of Input Enhancement is determined by the number, length and embedding dimension, which are set to 10, 10, and 768 in our setting. The size of the Tail Anchor is set to 100$\times$768. The total size of keys is (100+10)$\times$768. Therefore, the total number of trainable parameters amounts to 253,440. Compared to a ResNet-18 with 11,306,804 parameters, FedTA is efficient.

\noindent\textbf{Communication Cost.} Each client only needs to submit its own input enhancement and local prototypes to the server, with sizes of 76,800 and 768$\times$2 per class, respectively. Such small communication cost makes FedTA highly efficient, and also makes FedTA scalable for multi-clients.

\noindent\textbf{Privacy Protection.} For \textit{Input Enhancement}, on the one hand, ViT is frozen, and on the other hand, due to its minimal number of parameters, it contains extremely little information. Moreover, since this method does not use replay data to alleviate forgetting, privacy protection is further strengthened. However, the uploaded local prototypes are class-specific, and employing cross-mixing might easily reveal the original features, posing a certain degree of privacy risk. If we randomly mix Tail Anchor with features, this issue will be resolved.


\section{Conclusion}
This article extends the issue of data heterogeneity in static FL to the more realistic problem of spatial-temporal data heterogeneity in FCL. Empirical experiments are conducted to demonstrate that spatial-temporal data heterogeneity can cause parameter-forgetting and output-forgetting. Based on this finding, we first define the representative feature embedding of each class as the tail anchor. Then we propose FedTA by utilizing a frozen pre-trained ViT to mitigate parameter forgetting and combining Tail Anchors. Just as a ship in the ocean requires an anchor to hold its position, the features of samples also need a fixed location in the feature space to remain unaffected by the variations introduced by spatial-temporal data heterogeneity, thereby avoiding forgetting. Extensive experiments have verified the superiority of our method, and ablation studies demonstrate the effectiveness of each component, especially \textit{Tail Anchor}. Visualized results demonstrate that our method effectively fixes the features' relative positions, preventing them from being affected by spatial-temporal changes.

\section*{Acknowledgment}
This work was supported by the National Natural Science Foundation of china (No.~62476228), the Sichuan Science and Technology Program (No.~2024ZYD0180), the Graduate Representative
Achievement Cultivation Project of Southwestern University of Finance and Economics (Nos. JGS2024065, JGS2024004) and the Student Study Abroad Exchange Funding Program of Southwestern University of Finance and Economics.
\newpage
{
    \small
    \bibliographystyle{ieeenat_fullname}
    \bibliography{main}
}



\end{document}